\pdfoutput=1

\documentclass[11pt]{article}

\usepackage[preprint]{acl}

\usepackage{times}
\usepackage{latexsym}

\usepackage[T1]{fontenc}

\usepackage[utf8]{inputenc}

\usepackage{microtype}

\usepackage{inconsolata}

\usepackage{graphicx}

\usepackage{algorithm}
\usepackage{algorithmicx, algpseudocode}
\usepackage{graphicx}
\usepackage{subcaption}
\usepackage{amsmath}
\usepackage{amssymb}
\usepackage{amsfonts}
\usepackage{bm}
\usepackage{multirow}
\algrenewcommand\algorithmicrequire{\textbf{Input:}}
\algrenewcommand\algorithmicensure{\textbf{Output:}}
\allowdisplaybreaks

\title{Every Expert Matters: Towards Effective Knowledge Distillation for Mixture-of-Experts Language Models}

\author{Gyeongman Kim\textnormal{\textsuperscript{1}}$^*$ \quad Gyouk Chu\textnormal{\textsuperscript{1}}$^*$ \quad Eunho Yang\textnormal{\textsuperscript{1,2}} \vspace{0.03in}\\
  \textsuperscript{1}Korea Advanced Institute of Science and Technology (KAIST), South Korea \\ \textsuperscript{2}AITRICS, South Korea\\
  \texttt{\{gmkim, kyouwook, eunhoy\}@kaist.ac.kr}
}

\begin{document}
\maketitle
\def\thefootnote{*}\footnotetext{\,Equal Contribution.}\def\thefootnote{\arabic{footnote}}
\begin{abstract}
With the emergence of Mixture-of-Experts (MoE), the efficient scaling of model size has accelerated the development of large language models in recent years.
However, their high memory requirements prevent their use in resource-constrained environments. While knowledge distillation (KD) has been a proven method for model compression, its application to MoE teacher models remains underexplored.
Through our investigation, we discover that non-activated experts in MoE models possess valuable knowledge that benefits student models. We further demonstrate that existing KD methods are not optimal for compressing MoE models, as they fail to leverage this knowledge effectively.
To address this, we propose two intuitive MoE-specific KD methods for the first time: Knowledge Augmentation (KA) and Student-Aware Router (SAR), both designed to effectively extract knowledge from all experts.
Specifically, KA augments knowledge by sampling experts multiple times, while SAR uses all experts and adjusts the expert weights through router training to provide optimal knowledge.
Extensive experiments show that our methods outperform conventional KD methods, demonstrating their effectiveness for MoE teacher models.

\end{abstract}

\section{Introduction}


Mixture-of-Experts (MoE) architecture~\citep{jacobs1991adaptive, shazeer2017outrageously} is one of the major contributing factors to the rapid advancements of Large Language Models (LLMs)~\citep{jiang2024mixtral, team2024qwen1, liu2024deepseek}.
It allows the model to scale up while effectively improving the computational cost by utilizing only a subset of multiple experts during inference.
Despite the advantages afforded by MoE architectures in scaling model capacity, several limitations persist. One such challenge is that it requires significant GPU memory compared to the dense model due to a number of non-active parameters.
For this reason, the practical application of MoE models in resource-limited environments is generally challenging.
Hence, research into effectively compressing recent large-scale MoE models becomes imperative, particularly for deployment in resource-constrained environments. 


One of the notable compression techniques is knowledge distillation (KD)~\citep{hinton2015distilling}.
To facilitate student learning under teacher guidance, both the approach of using the teacher's output as supervised data~\citep{kim2016sequence, peng2023instruction, fu2023specializing} and the method to match the teacher's distribution with appropriate objective functions are widely adopted and actively researched.
Specifically, concerning the second method, many works have focused on designing suitable objective functions~\citep{wen2023f, ko2024distillm, agarwal2024generalized, wu2024rethinking} or on using student-generated output~\citep{lin2020autoregressive, gu2024minillm, agarwal2024generalized}.
Indeed, several models have successfully employed KD in practice, such as Phi~\citep{abdin2024phi} and Minitron~\citep{muralidharan2024compact, sreenivas2024llm}.

However, there has been no systematic development of KD methods specifically designed for the MoE teacher.
Recent KD studies have largely overlooked scenarios where the model to be compressed is based on the MoE structure.
While a few studies have applied KD to MoE teacher models~\citep{artetxe2021efficient, fedus2022switch, xue2022one}, they have used the conventional KD and have not thoroughly explored the effectiveness or challenges of distilling knowledge from MoE.
Therefore, these generalized approaches might not fully exploit the potential of MoE as a teacher.


In this paper, we introduce \textbf{MoE-specific knowledge distillation}, which can effectively distill knowledge from the MoE teacher.
To design such a specialized mechanism, we first conduct an in-depth analysis of MoE teacher during the basic KD process proposed by~\citet{sanh2019distilbert}.
We found that even non-selected experts have a significant amount of potentially useful knowledge, which remains unutilized. 
Inspired by this observation, we propose two different intuitive solutions for effectively extracting knowledge from all experts (see Figure \ref{fig3}). 
The first method, \textit{knowledge augmentation} (KA), employs sampling multiple times to decide which experts to activate based on their gate probabilities. Through this approach, a student can be provided with a variety of augmented knowledge from a single input data. 
The second method, \textit{student-aware router} (SAR), optimizes the router based on student feedback before distillation, enabling the router to determine optimal weights to aggregate knowledge from all experts. 


We apply our two approaches to Llama-MoE~\citep{zhu2024llama} models with five instruction datasets.
Considering the common practice of employing KD in memory-constrained settings, we utilize a dense student Sheared-Llama~\citep{xia2023sheared} rather than a MoE student.
The experimental results show that when the teacher model is MoE, our method consistently outperforms the existing KD baselines.
Additionally, the analysis of KA confirms that having a moderate amount of augmented knowledge is indeed beneficial.
Moreover, in SAR, we confirm that router updates in fact induce subtle changes in gate values, and these changes demonstrably enhance the performance of KD.

To summarize, our contributions are three-fold:
\begin{itemize}
    \item We empirically found that non-activated experts from MoE teacher also possess knowledge that is of great benefit to a student, yet remains unexploited by existing methods. 
    \item We propose two novel methods, knowledge augmentation (KA) and student-aware router (SAR), effectively utilizing the distributed knowledge from the entire experts. To the best of our knowledge, these are the first KD methods specifically designed for MoE teacher. 
    \item We evaluate our framework on 5 instruction-following datasets. The result shows that KA and SAR outperform the existing KD methods, underscoring the effectiveness and importance of leveraging the architectural characteristics of MoE teacher. 
\end{itemize}

\section{Related Works}

\paragraph{Knowledge distillation}
Knowledge distillation (KD)~\citep{hinton2015distilling} is a prevalent model compression technique, transferring knowledge from a large teacher model to a small student model.
Most of the early works focused on applying KD to the text classification tasks by imitating all the possible things of the teacher model, from output distribution~\citep{song2020lightpaff, liang2020mixkd} to hidden states~\citep{jiao2020tinybert, sun2019patient, park-etal-2021-distilling}, attention scores~\citep{wang2020minilm}, and so forth.
However, these methods relied on a fixed teacher that generates knowledge without being aware of the student’s learning characteristics, which often limited its effectiveness.
Thus, several methods are also devised to provide student-friendly knowledge~\citep{park2021learning, zhou2022bert, ren-etal-2023-tailoring}.

On the other hand, various studies are actively examining its application to text generation tasks.
The standard KD method minimizes the forward KL divergence between the output distributions of student and teacher at each time step~\citep{sanh2019distilbert} or directly trains the student with the generated text from the teacher~\citep{kim2016sequence, taori2023stanford, chiang2023vicuna, peng2023instruction}.
Recently, MiniLLM~\citep{gu2024minillm} explores a method to mix the distribution of the teacher with that of the student and use a policy gradient approach by optimizing the reverse KL divergence.
GKD~\citep{agarwal2024generalized} utilizes the student-generated on-policy data to receive feedback from the teacher with a generalized Jensen–Shannon (JS) divergence objective.
DistiLLM~\citep{ko2024distillm} applies skew KL divergence with their proposed adaptive off-policy mechanism.
Although these methods have shown remarkable results, all of the experiments have used dense models, and whether they also show good results for distilling the Mixture-of-Experts model has not yet been studied.

\paragraph{Mixture-of-Experts}
Mixture-of-Experts (MoE) \citep{shazeer2017outrageously, lepikhin2020gshard, fedus2022switch} is an efficient way to increase the model size by replacing the feed-forward network (FFN) with multiple experts and a gating network.
It dynamically activates different experts for each input token instead of using all parameters.
Since it has been known that MoE provides advantages including more efficient training~\citep{he2022fastermoe, gale2023megablocks} and faster inference than a dense model of the same size, many models such as Mixtral~\citep{jiang2024mixtral} and DeepseekMoE~\citep{dai2024deepseekmoe} have introduced MoE or its variants, demonstrating remarkably strong performance.
However, due to the disadvantage of high memory requirements, there have been some efforts to compress MoE into smaller dense models~\citep{artetxe2021efficient, fedus2022switch, xue2022one, guo2025deepseek}.
Nevertheless, they use the conventional KD~\citep{sanh2019distilbert} or train on the teacher's output sentence directly. To the best of our knowledge, there has been no attempt to develop the KD specifically optimized for MoE teacher.

\section{Preliminary}

\subsection{Knowledge Distillation}

KD minimizes the token-level distributional discrepancy between teacher and student. A standard approach to accomplish this minimization in the instruction-following setting is using the forward KL divergence~\citep{sanh2019distilbert}:
\begin{equation}
\mathcal{L}_{\text{KD}}=D_{KL}\big( p(\bm{y}|\bm{x}) \parallel q_\theta(\bm{y}|\bm{x}) \big),
\label{eqn:hintonkd}
\end{equation}
where $(\bm{x}, \bf{y})\in \mathcal{D}$, $\mathcal{D}$ denotes a dataset.
$\bm{x}$ and $\bm{y}$ represent the request and response, respectively, and this objective guides the student to learn by minimizing the distributional discrepancy in the only response part. $p$ and $q_\theta$ denote the probability distributions of the teacher and student, respectively.

Recently, MiniLLM~\citep{gu2024minillm} and GKD~\citep{agarwal2024generalized} suggest using reverse KL divergence and student-generated sequences to address the exposure bias problem. The objective reflecting these is as follows:
\begin{equation}
\mathcal{L}_{\text{student}}=D_{KL}\big( q_\theta(\bm{y}|\bm{x}) \parallel p(\bm{y}|\bm{x}) \big),
\label{eqn:student}
\end{equation}
where $(\bm{x}, \cdot)\in \mathcal{D}$ and $\bm{y} \sim q_\theta(\cdot|\bm{x})$.

\subsection{Mixture-of-Experts}

The sparse MoE layer consists of $N$ expert networks $\{E_{1}, \cdots, E_{N}\}$ and a router network $G$.
The router first computes the gate logits $H(x)\in \mathbb{R}^{N}$ for a single token representation $x$, which determines the likelihood of selecting each expert.
After normalization with a softmax function, top $k$ experts are selected based on this distribution, and their outputs are aggregated through a weighted sum.
In this work, we only focus on the noisy Top-$k$ gating introduced by~\citet{shazeer2017outrageously}.
This gating adds trainable Gaussian noise before Top-$k$ experts selection. The process can be described as follows:
\begin{align}
  \begin{split}
    H(x)_i = (x \cdot W_g)_i \ &+ \text{Standard}\text{Normal}() \ \cdot \\
    &\text{Softplus}((x \cdot W_{\text{noise}})_i),
\end{split} \label{eqn:noisytopk-logit} \\
  G(x) = \text{Softmax}&(\text{KeepTopK}(H(x), k)), \\
  y = \sum_{i=1}^N& G(x)_i E_i(x),
\end{align}
where $G(x)_{i}$ denotes the probability of $i$th experts being selected and \begin{equation}
\text{KeepTopK}(v, k)_{i}=
\begin{cases}
 v_{i} & \text{if } v_{i} \in \text{TopK}(v, k), \\
 -\infty & \text{otherwise}.
\end{cases} \nonumber
\end{equation}

The intrinsic characteristic of Top-$k$ routing may lead to a scenario where certain experts are always favored in the selection process.
In order not to negate the potential benefits of the MoE, distributing the workload across multiple experts to ensure their collective engagement is essential, which is called load balancing.
The noise term in $H(x)$ or the auxiliary loss as in Eq.~\eqref{eqn:LBL} helps prevent the model from always relying on the same few experts, allowing a more balanced distribution of workload among experts. 
The auxiliary loss~\citep{zhu2024llama} is as follows:
\begin{equation}
\label{eqn:LBL}
    \mathcal{L}_{\text{b}}=CV(\boldsymbol{m})^2+CV(\boldsymbol{P})^2,
\end{equation}
where $\boldsymbol{m} \in \mathbb{R}^{N}$ represents the set of token counts assigned to each of the $N$ experts within a batch, and $\boldsymbol{P} \in \mathbb{R}^{N}$ denotes the set of summed probabilities assigned to each expert in the batch. The coefficient of variation ($CV$) is defined as the ratio of the standard deviation $\sigma$ to the mean $\mu$, i.e., $CV(\boldsymbol{x}) = \sigma(\boldsymbol{x})/\mu(\boldsymbol{x})$. Minimizing this encourages a more uniform distribution, which is desirable for balancing the expert load. 

\section{Method}

\subsection{Motivation}
\label{approach:motivation}

\begin{figure*}[t!]
  \centering
    \subfloat[Llama-MoE-3.5B (4/16)]
    {\label{fig1:subfig-a}
    \includegraphics[width=0.32\linewidth]{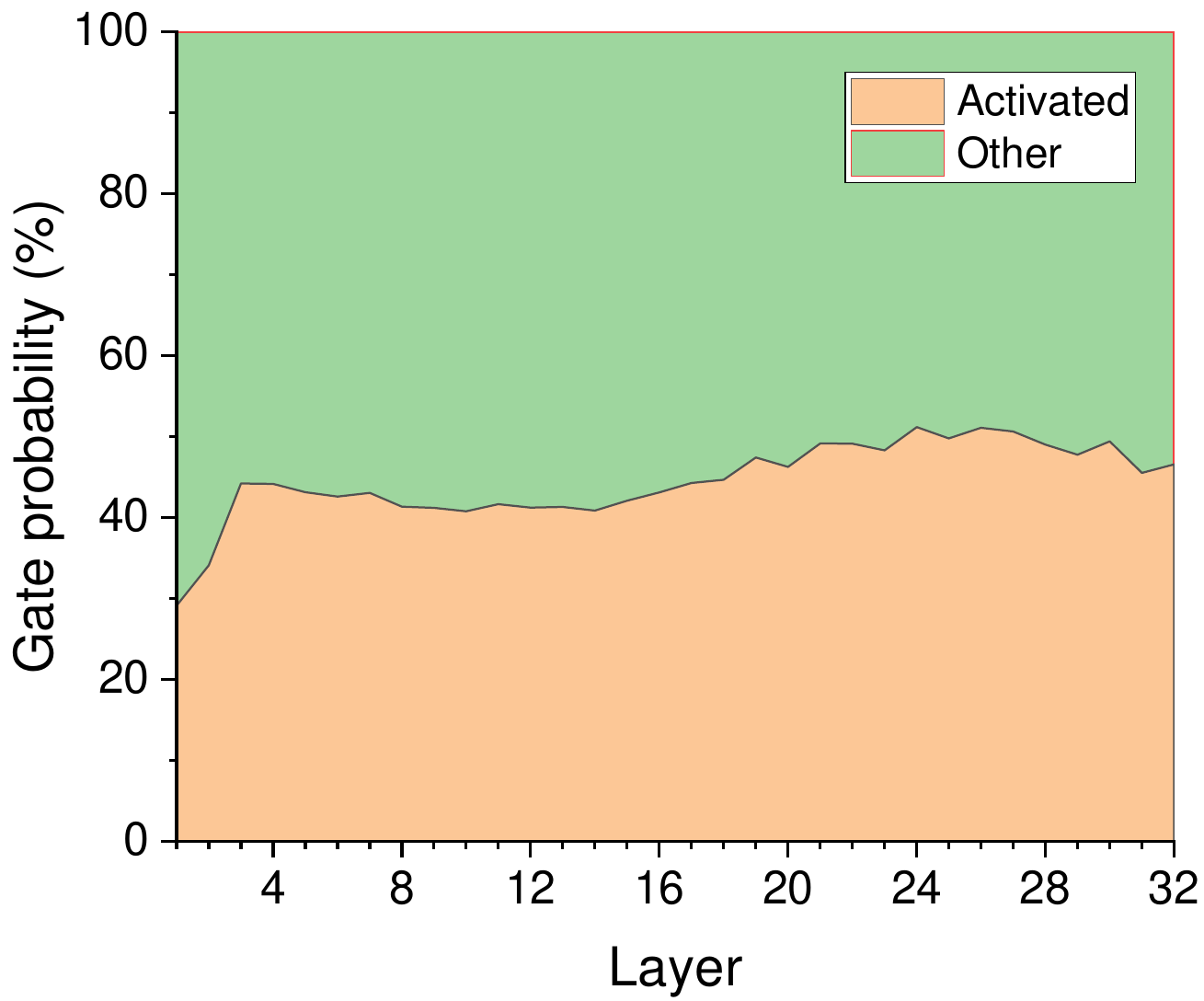}
    }
    \hfill
    \subfloat[Llama-MoE-3.5B (2/8)]
    {\label{fig1:subfig-b}
    \includegraphics[width=0.32\linewidth]{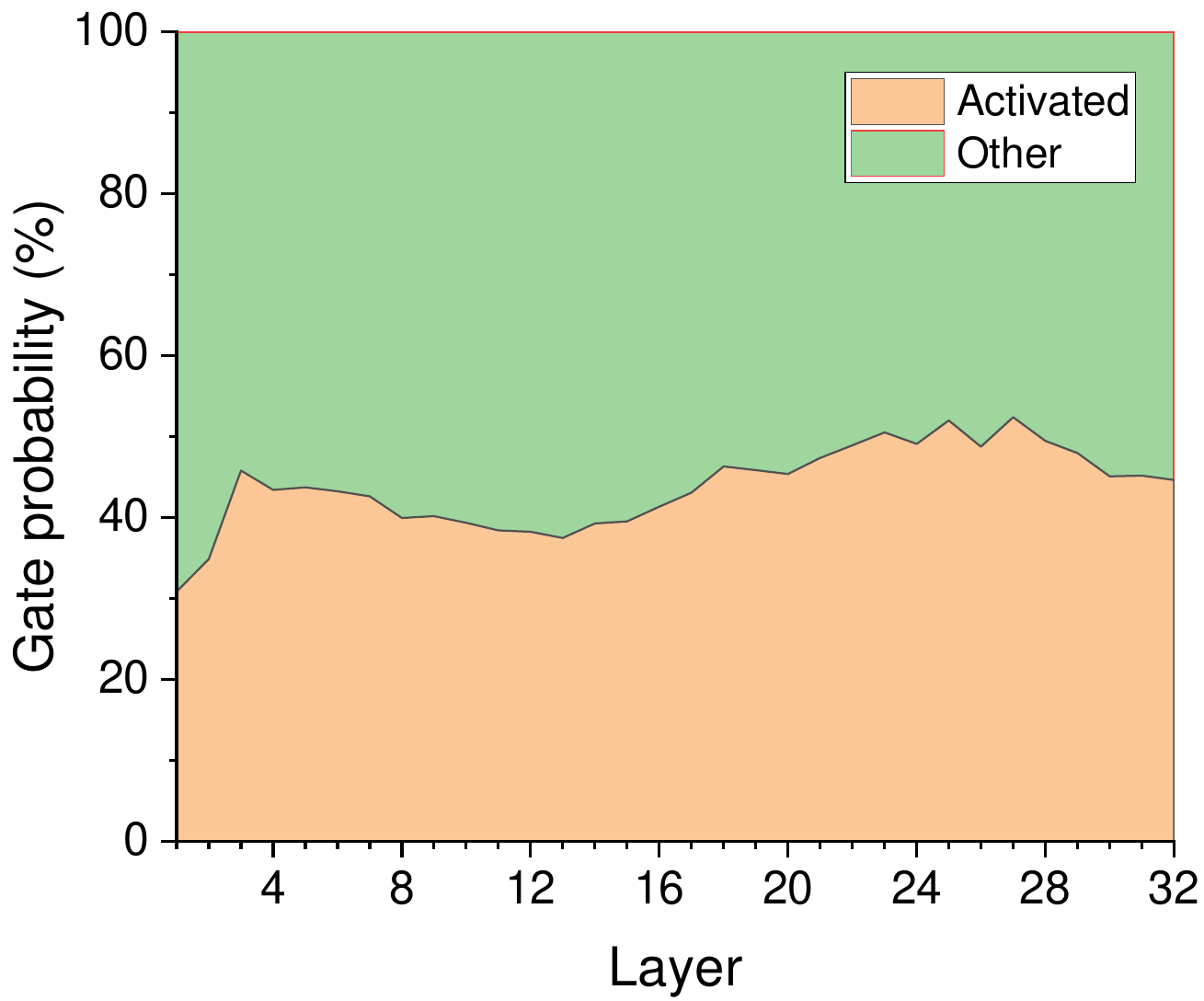}
    }
    \hfill
    \subfloat[Llama-MoE-3.0B (2/16)]
    {\label{fig1:subfig-c}
    \includegraphics[width=0.32\linewidth]{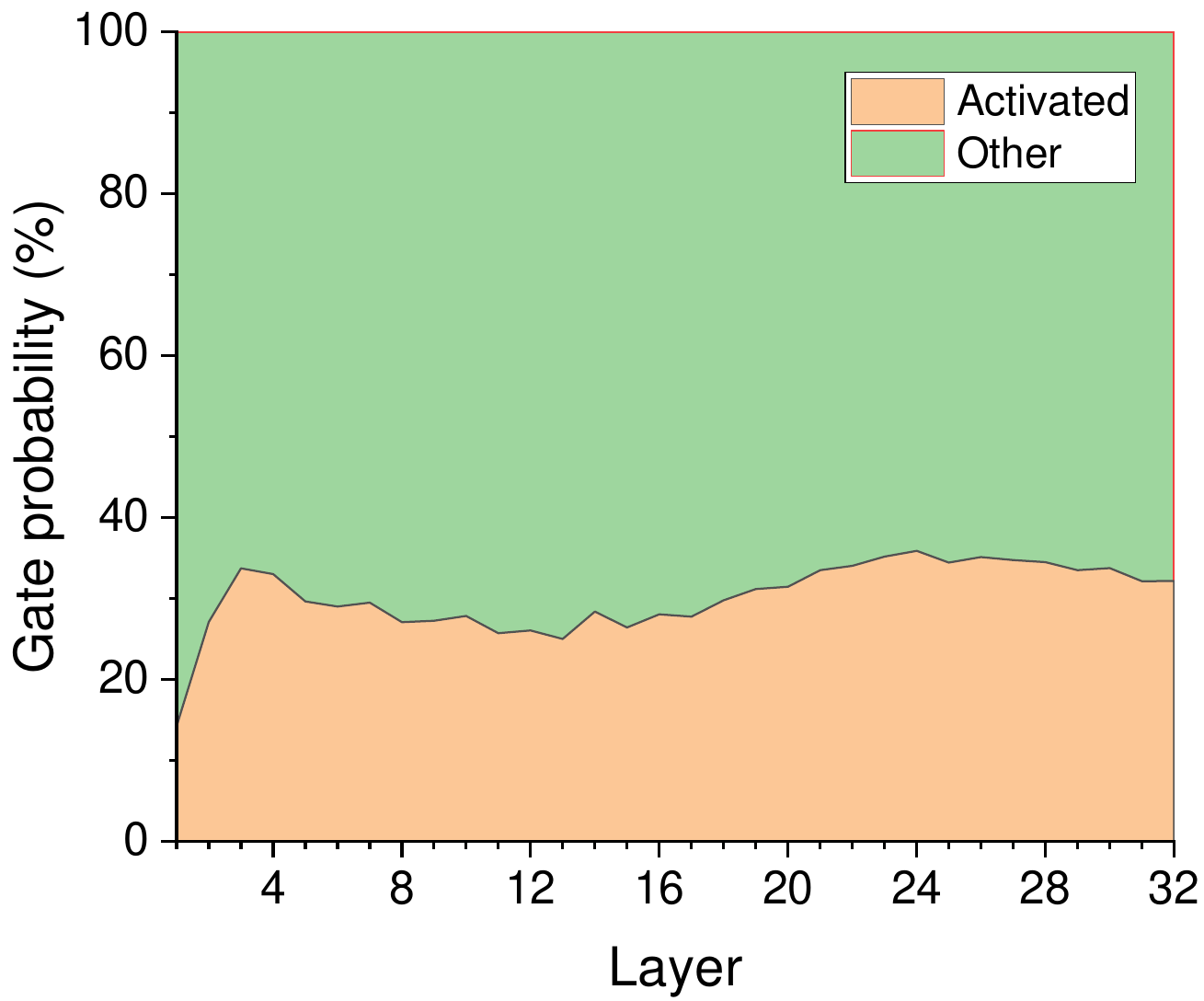}
    }
  \caption{Sum of the gate probabilities for activated and non-activated experts per layer during distillation. The $(k/N)$ after each model name indicates that $k$ out of $N$ experts are activated. Across most layers of all Llama-MoE models, the sum of gated probabilities of activated experts is less than 50\%.}
  \label{fig1}

\end{figure*}


To investigate how the MoE teacher distills the knowledge during the classical KD process, we first analyze the distribution of gate probabilities. 
The gate probability refers to the normalized values of the gate logits $H$ through the softmax function. The Top-$k$ experts are selected based on these gate logits, and gate logits are also used to compute the weights during the weighted summation of expert outputs. Therefore, the gate probability can be considered an indicator of how useful each expert is.
In this section, we use Llama-MoE~\citep{zhu2024llama} models as teachers and do the conventional KD~\citep{sanh2019distilbert} into Sheared-Llama~\citep{xia2023sheared} model which is a dense model. The training data is a subset of Dolly~\citep{DatabricksBlog2023DollyV2}, and we evaluate our model on five instruction datasets. For further details, please see the Section~\ref{experiments:setup}.

Figure~\ref{fig1} presents a visualization of the average of the sum of gate probabilities for used experts and that for unused experts in each layer across all training data during distillation. As shown in Figure~\ref{fig1}, the sum of gate probabilities for the group of activated experts never exceeds 50\% in most of the layers of all models.
Although this may be an effect of the auxiliary loss for load balancing, considering that gate probability reflects how useful an expert is, 
it implies that a significant portion of potentially valuable knowledge from non-activated experts is not being leveraged. 
Thus, effective extraction and utilization of this unexploited knowledge could bring additional benefits to the student model during the distillation process, as more diverse and complementary knowledge would be incorporated into the learning. 

\begin{figure}[t]
  \subfloat[Distilled student under MoE teacher]
  {\label{fig2:subfig-a}
  \includegraphics[width=\columnwidth]{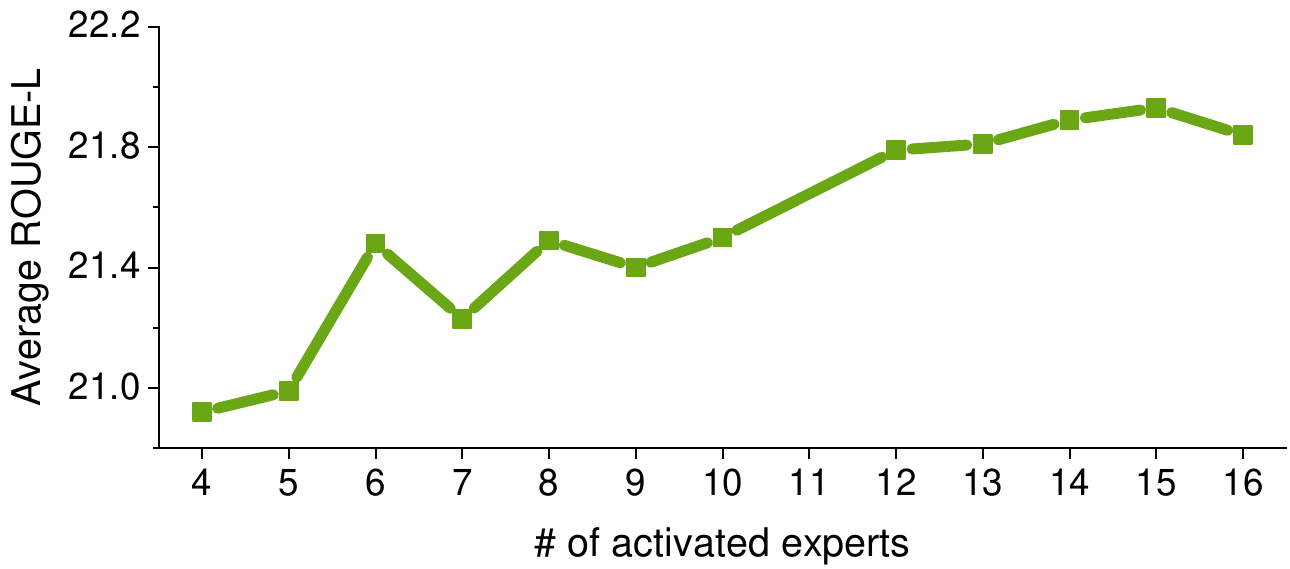}
  }
  \vspace{3mm}
  \subfloat[MoE teacher]
  {\label{fig2:subfig-b}
  \includegraphics[width=\columnwidth]{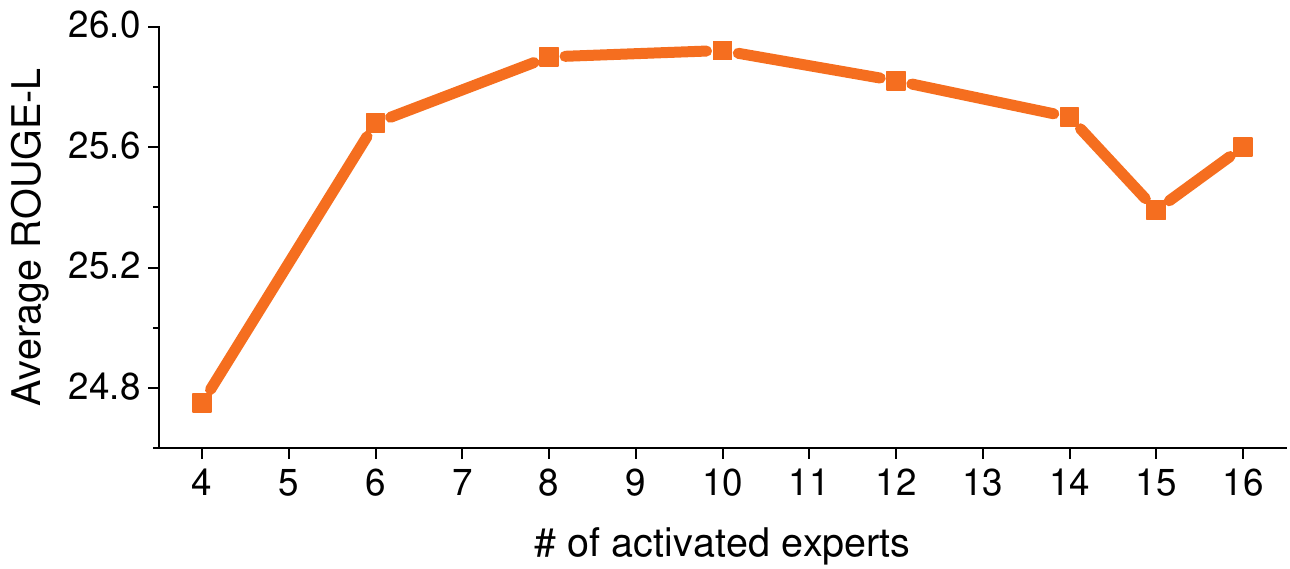}
  }
  \caption{Performance of the MoE teacher model and the student model after distillation with varying numbers of utilized experts $k$ (originally 4). As $k$ increases, the effectiveness of distillation improves, leading to better student performance. However, the performance of the teacher model itself does not necessarily improve with a larger $k$.}
  \label{fig2}
  \vspace{-0.2in}
\end{figure}

\begin{figure*}[t]
  \centering
    \subfloat[Knowledge Augmentation (KA)]
    {\label{fig3:subfig-b}
    \includegraphics[width=0.48\linewidth]{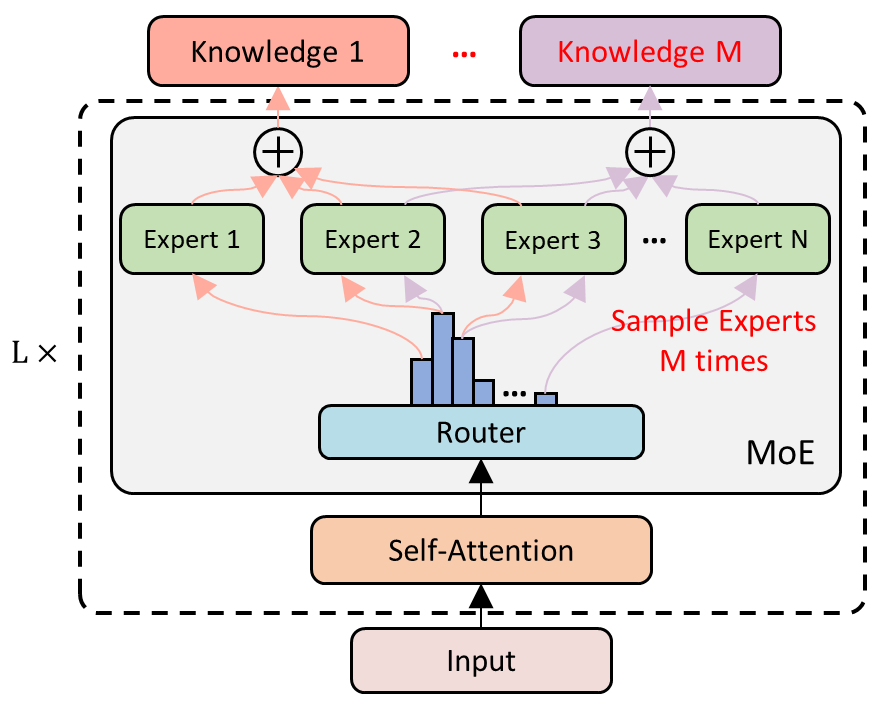}
    }
    \hfill
    \subfloat[Student-Aware Router (SAR)]
    {\label{fig3:subfig-c}
    \includegraphics[width=0.48\linewidth]{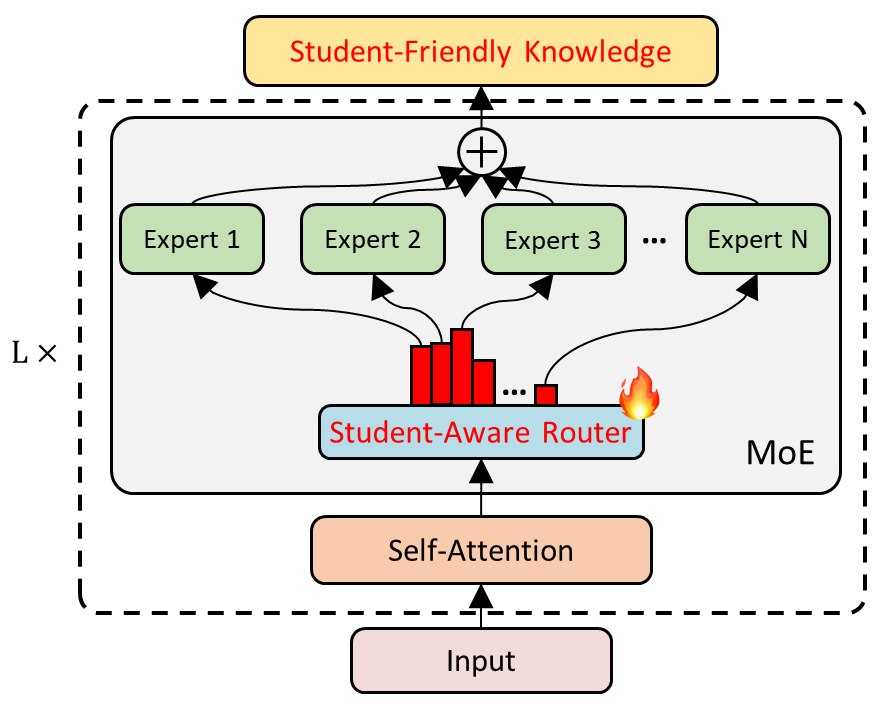}
    }
  \caption{An overview of our proposed KD methods specifically designed for the MoE. In knowledge augmentation, we either select the $\text{top }N-1$ experts or sample $N-1$ experts based on the gate probability. We do this $M$ times to augment various knowledge. In student-aware router, we train the router network with student feedback before the distillation. It enables the router to determine the optimal weights, thereby facilitating the student's acquisition of knowledge from all experts.}
  \label{fig3}
\end{figure*}

To reflect this observation, we simply increase the number of selected experts $k$ during the distillation process.
Using the Llama-MoE-3.5B (4/16) model as the teacher model, we perform knowledge distillation by gradually increasing $k$ from 4 to 16 and evaluate the performance of both the teacher and student models.
The model performance is measured by the average ROUGE-L scores across five instruction-following datasets (Section~\ref{experiments:setup} for more details).
The results are shown in Figure~\ref{fig2}.

Based on the results, we observe that using more experts does not necessarily increase the performance of the teacher, but it certainly increases the performance of the student, except when all are used.
This suggests that the improvement in the student’s performance is not directly due to the teacher’s performance enhancement. Nevertheless, we observe that using most of the non-activated experts proves to be practically beneficial for the student, and this implies that non-activated experts hold valuable knowledge. The reason for this could be that during the MoE training process, due to load balancing, different sets of experts are activated for the same input data, causing the knowledge to be distributed across multiple experts. However, conventional KD typically relies on using only the Top-k experts, which fails to account for this.

Therefore, the core challenge in knowledge distillation for MoE teacher lies in \textbf{effectively extracting and transferring the knowledge that is distributed across all experts} to empower student learning. Successfully addressing this challenge is the key to fully leveraging the architectural characteristics of MoE teacher models in guiding student models. 

\subsection{Knowledge Augmentation}
\label{KA}

\begin{algorithm}[t]
\caption{: Knowledge Augmentation (KA)}
\label{alg:KA}
\begin{algorithmic}
\Require student model $q_\theta$, data distribution $p_x$, number of teacher forward $M$, training step $K$, learning rate $\eta$
\For{each step $k=1,...,K$}
    \State Sample a request $\bm{x}$ from $p_x$
    \State Sample a response $\bm{y}$ from $q_\theta(\cdot|\bm{x})$
    \For{each step $m=1,...,M$}
        \State Update $\theta \leftarrow \theta-\eta \nabla \mathcal{L}_{\text{student}}$  \quad $\triangleright$ Eq.~\eqref{eqn:student}
    \EndFor
\EndFor
\State \Return $\theta$
\end{algorithmic}
\end{algorithm}

The first method to effectively utilize distributed knowledge across all experts is the knowledge augmentation (KA).
Following the previous observation, we use $N-1$ experts for each layer where $N$ is the total number of experts.
Specifically, in each MoE layer, $N-1$ experts are selected by sampling from a gate probability distribution with probability $\lambda$.
Therefore, by selecting the Top $N-1$ experts with probability $1-\lambda$, we can consistently generate knowledge that is similar to the Top-k selection while incorporating slightly different knowledge.
This strategy allows the augmentation of diverse knowledge and balances the trade-off between consistency and diversity of knowledge with parameter $\lambda$.
The formulation of KA is as follows:
\begin{align}
&\mathbf{E} =
\begin{cases}
 \text{Sampled } N-1 \ \text{experts} & \text{w.p. } \lambda, \\
 \text{Top } N-1 \ \text{experts} & \text{w.p. } 1-\lambda,
\end{cases} \nonumber\\[5pt]
&\text{KA}(v, \mathbf{E})_{i} =
\begin{cases}
 v_{i} & \text{if } (i \text{th expert}) \in \mathbf{E}, \\
 -\infty & \text{otherwise},
\end{cases} \\[5pt]
&G^{\text{KA}}(x)= \ \text{Softmax}(\text{KA}(H(x), \mathbf{E})),
\end{align}
where $\mathbf{E}$ denotes the set of selected experts.

In each iteration, the teacher is forwarded $M$ times for the same input using the KA method, augmenting $M$ pieces of knowledge, which are transferred to the student.
Following GKD~\citep{agarwal2024generalized}, the response part $\bm{y}$ of the input is generated by the student, treating it as a pseudo-target, to mitigate exposure bias~\citep{arora2022exposure}.
Furthermore, the student's learning objective is the reverse KL divergence.
We summarize the entire KA procedure in Algorithm~\ref{alg:KA}.


\subsection{Student-Aware Router}
\label{SAR}

The second method is the student-aware router (SAR). Instead of merely selecting which experts to use, SAR takes a step further by directly optimizing the router to achieve an optimal weighted sum across all expert outputs. Inspired by the concept of student-friendly knowledge distillation, SAR updates the teacher’s router using student feedback, ensuring that the generated knowledge is more useful to the student.

SAR undergoes two stages in each iteration: router update and knowledge distillation. First, the router weights, $W_g$ and $W_{\text{noise}}$ in Eq.~\eqref{eqn:noisytopk-logit}, are optimized using student feedback~\citep{kim-etal-2024-promptkd} and auxiliary loss for load balancing. Only the router components of the MoE teacher are updated, while all other parameters remain fixed. After updating the router, the modified router is used to generate knowledge, which is then distilled into the student. At this stage, all experts are activated, and their outputs are aggregated through a weighted sum based on the modified router.

Similar to KA, SAR also uses pseudo-targets generated by the student and trains the router using reverse KL divergence: \begin{align}
    \mathcal{L}_{\text{SAR}}=&D_{KL}\big( p(\bm{y}|\bm{x}) \parallel q_\theta(\bm{y}|\bm{x}) \big) + \beta\mathcal{L}_{\text{b}}.
    \label{eqn:SAR}
\end{align} 
Here, $\beta$ is a coefficient for the auxiliary loss, which is set to 0.01 following the teacher model~\citep{zhu2024llama}. 
The entire SAR process is summarized in Algorithm~\ref{alg:SAR}.

\section{Experiments}

\subsection{Experimental Setup}
\label{experiments:setup}

\paragraph{Settings} 
Following~\citet{gu2024minillm}, \texttt{databricks -dolly-15k}\,~\citep{DatabricksBlog2023DollyV2} is partitioned into 14k samples for the training set, 500 samples for the validation and test sets, respectively. In addition to the test set of Dolly, we evaluate 4 extra instruction-following datasets: SelfInst~\citep{wang-etal-2023-self-instruct}, 252 user-oriented instruction-following samples, Vicuna~\citep{chiang2023vicuna}, 80 questions used in the Vicuna evaluation, S-NI, 9k samples from the test set of \textsc{Super-NaturalInstructions}~\citep{wang2022super}, and UnNI, randomly sampled 10k samples from the core set of \textsc{UnnaturalInstructions}~\citep{honovich-etal-2023-unnatural}. We adopt the ROUGE-L~\citep{lin2004rouge} score as the evaluation metric.

\begin{algorithm}[t]
\caption{: Student-Aware Router (SAR)}
\label{alg:SAR}
\begin{algorithmic}
\Require student model $q_\theta$, data distribution $p_x$, teacher's router $W_{g}$ and $W_{\text{noise}}$, training step $K$, learning rate $\eta$

\For{each step $k=1,...,K$}
\State Sample a request $\bm{x}$ from $p_x$
\State Sample a response $\bm{y}$ from $q_\theta(\cdot|\bm{x})$
\State Update $W_{g} \leftarrow W_{g}-\eta \nabla \mathcal{L}_{\text{SAR}}$  \quad $\triangleright$ Eq.~\eqref{eqn:SAR}
\State Update $W_{\text{noise}} \leftarrow W_{\text{noise}}-\eta \nabla \mathcal{L}_{\text{SAR}}$
\State Update $\theta \leftarrow \theta-\eta \nabla \mathcal{L}_{\text{student}}$  \quad $\triangleright$ Eq.~\eqref{eqn:student}
\EndFor
\State \Return $\theta$
\end{algorithmic}
\end{algorithm}

\begin{table*}[t]
\centering
\begin{tabular}{llcccccc}
\hline
\multirow{2}{*}{\shortstack{Model\\(Teacher $\rightarrow$ Student)}} & \multirow{2}{*}{Method} & \multicolumn{5}{c}{Instruction-following datasets} & \multirow{2}{*}{Average} \\ 
\cline{3-7} & & Dolly & SelfInst & Vicuna & S-NI & UnNI \\ 
\hline
\hline
Llama-MoE-3.5B (4/16) & SFT & 26.20 & 18.61 & 16.88 & 30.29 & 31.79 & 24.75 \\
Llama-MoE-3.5B (2/8) & SFT & 26.39 & 16.97 & 17.20 & 30.40 & 32.81 & 24.76 \\
Llama-MoE-3.0B (2/16) & SFT & 26.35 & 17.64 & 16.86 & 27.59 & 30.42 & 23.77 \\
Sheared-Llama-2.7B & SFT & 26.07 & 18.55 & 17.50 & 27.64 & 31.13 & 24.18 \\
Sheared-Llama-1.3B & SFT & 23.83 & 14.82 & 15.93 & 26.33 & 28.21 & 21.82 \\
\hline
\multirow{2}{*}{\shortstack{Sheared-Llama-2.7B\\$\rightarrow$ Sheared-Llama-1.3B}} 
 & KD & 24.68 & 13.44 & 16.16 & 26.37 & 29.09 & 21.95\\ 
& GKD & \textbf{26.36} & \textbf{16.67} & \textbf{18.20} & \textbf{29.09} & \textbf{34.12} & \textbf{24.89} \\
\hline
\multirow{5}{*}{\shortstack{Llama-MoE-3.5B (4/16)\\$\rightarrow$ Sheared-Llama-1.3B}} 
 & KD & 23.58 & 13.82 & 15.25 & 24.59 & 27.37 & 20.92\\
 & GKD & 25.86 & 16.72 & \textbf{18.61} & 29.61 & 34.55 & 25.07\\
& ALL (Ours)& 26.03 & 16.98 & 18.59 & 30.13 & 34.88 & 25.32 \\
& KA (Ours)& \textbf{26.58} & 16.98 & 18.38 & 30.51 & \textbf{36.11} & 25.71 \\
& SAR (Ours)& 26.32 & \textbf{18.24} & 18.06 & \textbf{31.88} & 35.05 & \textbf{25.91}\\
\hline
\multirow{5}{*}{\shortstack{Llama-MoE-3.5B (2/8)\\$\rightarrow$ Sheared-Llama-1.3B}} 
 & KD & 23.07 & 13.92 & 15.29 & 24.87 & 27.40 & 20.91\\
 & GKD & 25.64 & 15.54 & 18.29 & 29.11 & 32.80 & 24.28\\
& ALL (Ours)& \textbf{26.40} & 16.78 & \textbf{18.45} & 28.68 & 33.57 & 24.78 \\
& KA (Ours)& 26.32 & 17.30 & 17.11 & \textbf{32.49} & \textbf{37.58} & \textbf{26.16} \\
& SAR (Ours)& 26.30 & \textbf{18.31} & 17.11 & 31.47 & 35.00 & 25.64\\
\hline
\multirow{5}{*}{\shortstack{Llama-MoE-3.0B (2/16)\\$\rightarrow$ Sheared-Llama-1.3B}} 
 & KD & 23.20 & 13.51 & 15.01 & 23.85 & 26.92 & 20.50\\
 & GKD & 25.43 & 16.43 & \textbf{18.52} & 28.15 & 34.71 & 24.65\\
& ALL (Ours)& 25.99 & 15.05 & 18.06 & 29.15 & 33.55 & 24.36 \\
& KA (Ours)& \textbf{26.06} & 16.18 & 18.30 & 30.10 & \textbf{35.92} & 25.31 \\
& SAR (Ours)& 25.87 & \textbf{17.39} & 17.84 & \textbf{31.20} & 34.92 & \textbf{25.44}\\
\hline
\end{tabular}
\caption{Evaluation results on five instruction-following datasets and their average, assessed using the ROUGE-L metric. Each reported score represents the average across five distinct random seeds. The best score for each case is highlighted in \textbf{boldface}.}
\label{tab:main_result}
\end{table*}

\paragraph{Models}
To verify the effectiveness of proposed KD methods tailored for MoE, we need to compare the performance of KD from dense to dense with that from MoE to dense.
For this comparison to be fair, dense teacher and MoE teacher should have comparable performances.
This ensures that any performance improvements can be directly ascribed to the proposed method rather than the teacher's inherent capability.
Additionally, the tokenizers of both models must be the same to compare token-level distributions.

To satisfy the above critical conditions, we use three Llama-MoE~\citep{zhu2024llama} models as the MoE teachers, Sheared-Llama~\citep{xia2023sheared} 2.7B as the dense teacher, and Sheared-Llama 1.3B as the dense student. Sheared-Llama 2.7B  exhibits comparable performance to Llama-MoE model, with a lower number of activated parameters.
Both teacher models and the student model were initially fine-tuned with the Dolly training set before knowledge distillation, following the previous works~\citep{agarwal2024generalized, gu2024minillm}.

\paragraph{Baseline}
We compare our two approaches with three baselines: (1) supervised fine-tuning (SFT) directly fine-tunes the model on golden responses, which does not involve knowledge distillation; (2) KD~\citep{sanh2019distilbert} uses the teacher's distribution with forward KL divergence; (3) GKD~\citep{agarwal2024generalized} uses the mixture of fixed data and on-policy student-generated outputs.
Despite recent advancements and variants, GKD remains a representative study utilizing KL divergence or its variants and student-generated outputs, making it a suitable baseline for our experiment.
Based on their reported performance, GKD computes reverse KL divergence with only student-generated outputs in this paper.
For our methods, we set a sampling ratio $\lambda=0.05$ and the number of augmented samples $M=2$ in the KA method.
To validate our observation on the MoE teacher, we exclude the router update stage from SAR and simply activate all experts. This approach is referred to as ALL.
Further details on the experimental setup are summarized in the Appendix~\ref{sec:appendix-setup-detail}.

\subsection{Results}
\label{experiments:results}

We present the results of KA and SAR with baselines on 5 datasets in Table~\ref{tab:main_result}.

First, when comparing the SFT results of three Llama-MoE models, the performance is better when there are more activated experts with the same total number of experts. 
Also, if the total activated parameters are similar, the performance is also comparable. 
Note that the dense teacher Sheared-Llama-2.7B indeed shows a similar performance compared to MoE teachers.

Second, we compare the performance between dense and MoE teachers for the two baselines, KD and GKD.
Surprisingly, despite the MoE teacher having performance that is similar to or even slightly better than the dense teacher, 
both methods demonstrate that the dense model serves as a better teacher for the student. 
For KD, the student trained by the dense teacher always outperforms the student trained by the MoE teachers. 
This holds true under GKD as well, except for the Llama-MoE-3.5B (4/16) case. 
These results highlight that the existing KD methods are not optimized for extracting knowledge from the MoE teacher. 

Third, our proposed methods, knowledge augmentation and student-aware router, achieve higher performance than baselines when the teacher model is MoE.
This result highlights that both methods are specifically designed for the MoE teacher.
Therefore, when the teacher model is MoE, it is important to carefully consider the architectural characteristics of MoE and effectively extract knowledge that is distributed across all experts.


Lastly, the ALL approach, which simply activates all experts, outperforms the baselines in most cases but falls short of our proposed methods.
This result aligns with the observation in Section~\ref{approach:motivation}, suggesting that while non-activated experts contain useful knowledge, simply utilizing all of them may not be the optimal strategy.
Furthermore, the comparison with SAR demonstrates the effectiveness of router updates.

The qualitative results of our methods and the baselines are summarized in Appendix~\ref{sec:qualitative_results}, demonstrating that our methods produce responses most closely resembling the ground truth.

\begin{figure}[t]
  \includegraphics[width=0.93\linewidth]{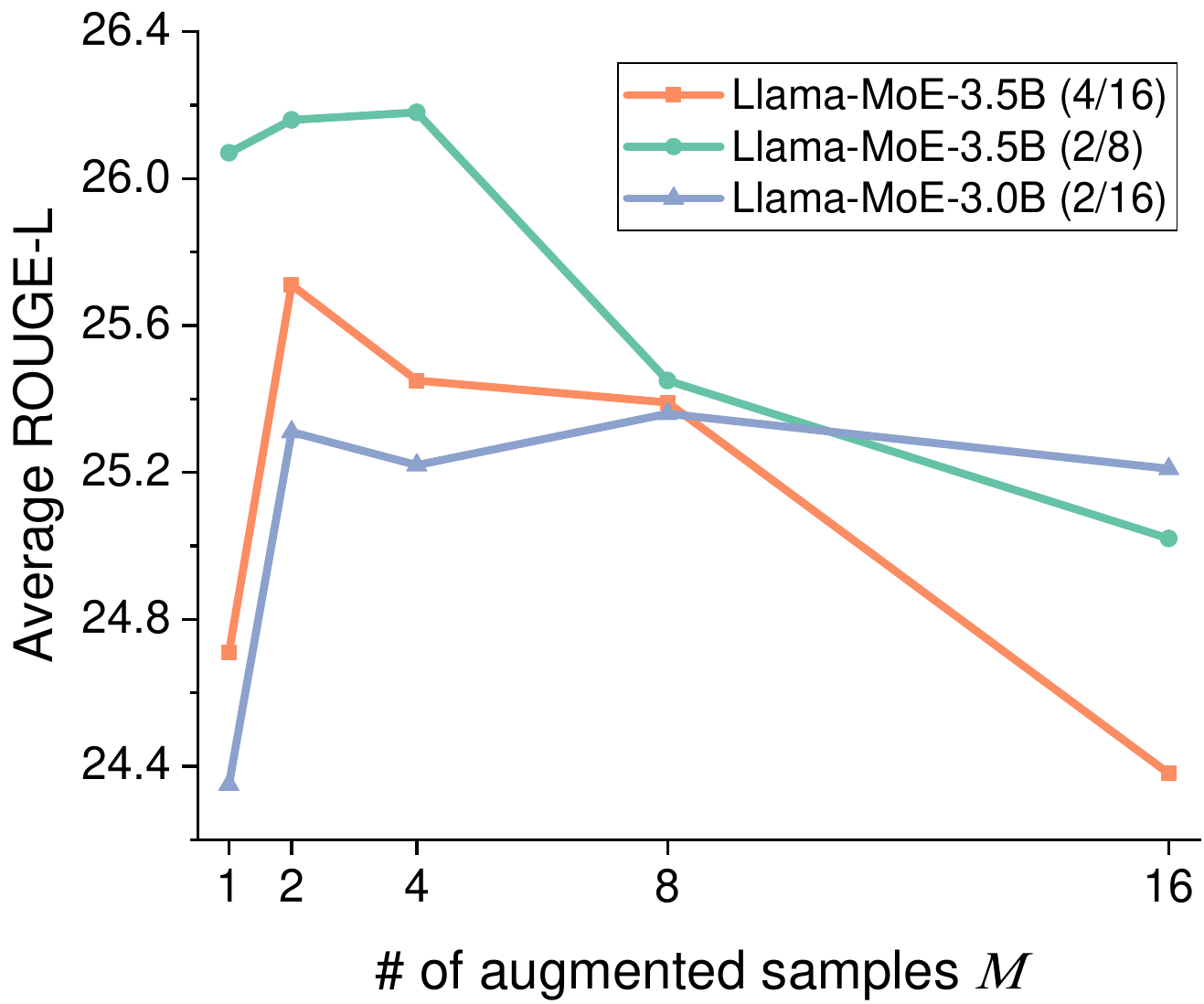}
  \caption{Average performance of KA for a different number of samples, $M$, across all test data. $\lambda$ is fixed at 0.05. For each MoE teacher, the best performing $M$ differs. If $M$ is too large, all models exhibit reduced performance.}
  \label{fig:effect_of_M}
  \vspace{-0.1in}
\end{figure}

\subsection{Analysis}
\paragraph{Hyperparameters in KA}
We ablate various values of $M$, the number of augmented samples in KA.
Figure~\ref{fig:effect_of_M} shows the performance for different numbers of samples, $M$.
It indicates that the optimal $M$ value varied across different models.
Nevertheless, the appropriate value of $M$ generally leads to beneficial augmentation.
However, when $M$ is excessively large, performance consistently degrades across all models.
This is because too large values can lead to the generation of overly diverse knowledge for identical input due to the inherent randomness of sampling.
Consequently, such excessive diversity can be detrimental to the overall performance, as it may introduce nonsense or unhelpful knowledge.

We also ablate various values of $\lambda$, the probability of randomly sampling experts. The results are in Appendix~\ref{sec:appendix-KA-lambda}.

\paragraph{Shift of gate probability in SAR}

In Table~\ref{tab:main_result}, we compared the results of ALL and SAR and verified that training the routers of MoE teacher is indeed helpful.
For a more rigorous analysis, we examine the changes in the gate probability distribution that occurred as the router network learned to be more student-aware.

\begin{figure}[t]
  \includegraphics[width=0.96\linewidth]{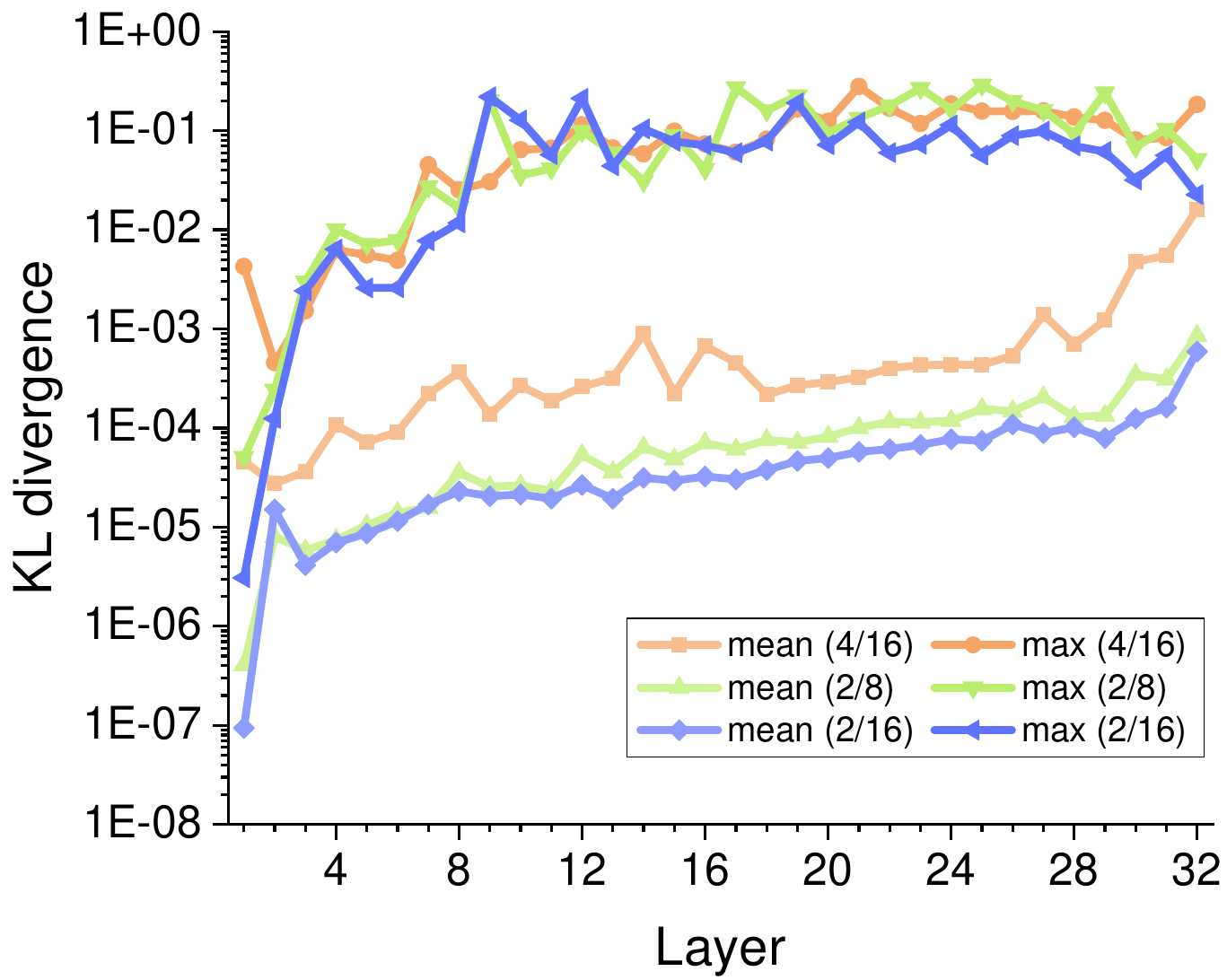}
  \caption{KL divergence of gate probabilities between original router and router trained with SAR method. The mean value is averaged over all tokens in training data. Consistently, KL divergence increases with layer depth.}
  \label{fig:sar_analysis}
  \vspace{-0.1in}
\end{figure}

Figure~\ref{fig:sar_analysis} presents the layer-wise KL divergence of gate probabilities between the original teacher MoE and the teacher whose routers are trained with SAR. For all tokens of the training data, the maximum and average values are shown.
For every teacher model, KL divergence increases with greater layer depth.
The reason is that by learning the router in a student-friendly way, the modified gate probability affects the representation of the layer immediately following.
This effect accumulates so that later layers have more different gate probabilities than the existing router.
Eventually, these changes in gate probability have led to a more effective knowledge delivery to the student.

\section{Conclusion}
In this paper, we first observe that non-activated experts in MoE teachers contain valuable knowledge that can benefit the student model.
Based on this observation, we assert that existing KD methods are suboptimal for distilling MoE models, as they do not fully utilize all experts.
To address this issue, we propose two MoE-specific KD methods for the first time: knowledge augmentation and student-aware router.
Our experimental results show that our methods outperform the baseline, clearly demonstrating the effectiveness of our approach in leveraging the full potential of MoE teacher models.

\section*{Limitations}
We acknowledge the limitations arising from the rigorous experimental conditions. In addition to the common yet imperfect situation where teacher and student must use the same tokenizer, dense teacher and MoE teacher should have comparable performances.
This condition is necessary to show that our method is an effective KD specialized for MoE.
However, it is difficult to find a setup that satisfies these conditions other than the setting that we used in our experiment (Llama-MoE~\citep{zhu2024llama} for the teacher and Sheared-Llama~\citep{xia2023sheared} for the student).
We leave this for future work to explore, in conjunction with either emerging new methods~\citep{boizard2024towards, zhang2024dual} or by combining our method with existing ways~\citep{xue2022one}.


\bibliography{custom}

\appendix

\section{Experimental Setup Details}
\label{sec:appendix-setup-detail}

For training, we utilize the AdamW optimizer~\citep{loshchilov2017decoupled} with a batch size of 16. The learning rates for both the router and student models are set to 1e-5, and training is conducted for 10 epochs. The training and generation processes are conducted with a maximum sequence length of 512 and a maximum request length of 256. During generation, we apply top-k and top-p sampling with values of 0 and 1.0, respectively, while maintaining a fixed temperature of 1.0.
All experiments in this study are conducted on 4 Intel Gaudi v2 accelerators using SynapseAI 1.18.0.

To ensure consistency in instruction-following tasks, all datasets are pre-processed by converting instruction-response pairs into a standardized sentence structure, following the approach used in previous studies~\citep{gu2024minillm}. Model evaluation is performed using the ROUGE-L score~\citep{lin2004rouge}, which has been shown to correlate well with human preferences in instruction-following assessments~\citep{wang2022super}. The best-performing checkpoint on the validation set, determined by the ROUGE-L score, is selected for final evaluation. All evaluations are performed across five different random seeds, and the reported results reflect the average performance.

\section{Effects of $\lambda$ in KA}
\label{sec:appendix-KA-lambda}

\begin{table}[h]
\centering
\resizebox{\columnwidth}{!}{%
\begin{tabular}{c|cccccc}
\hline
$\lambda$ & Dolly & SelfInst & Vicuna & S-NI & UnNI & Average \\
\hline 				
0.01 & 26.23 & 17.55 & 18.10 & 31.19 & 35.52 & 25.72 \\
0.03& 26.35 & 17.15 & 18.65 & 29.90 & 34.68 & 25.35 \\
0.05& 26.58 & 16.98 & 18.38 & 30.51 & 36.11 & 25.71 \\
0.1 & 26.34 & 16.32 & 18.96 & 30.99 & 36.01 & 25.73 \\
0.2 & 26.49 & 16.24 & 18.88 & 29.99 & 34.08 & 25.14 \\
\hline
\end{tabular}
}
\caption{Instruction-following performance of KA across all test data for different values of $\lambda$, which represents the probability of sampling experts. Here, we only use Llama-MoE-3.5B (4/16) model and set $M=2$.}
\label{tab:ablation_lambda}
\vspace{-0.1in}
\end{table}

\begin{table*}[!t]
\centering
\begin{tabular}{l|p{0.8\linewidth}}
\hline
\multicolumn{2}{c}{Case 1} \\
\hline
Instruction & Convert a disfluent question to a proper question. A disfluent question is a question that has some interruptions in it while framing. A proper question is the correct form of the question without any disfluency. \\
\hline
Input & What do no I meant Who do some tax agents want to make an impression on during an arrest? \\
\hline
Ground Truth & Who do some tax agents want to make an impression on during an arrest? \\
\hline
\hline
SFT & Tax agents want to make an impression on during an arrest. \\
\hline
KD & An arrest is usually a life changing event. Ensure that the police officer Who makes the arrest understands that you do not do it on a conscious level. Use the hook offered to manually fill in the blank to make sure that the question sounds proper. \\
\hline
GKD & They want to make an impression on during an arrest that they will be able to collect from the tax payer on the charge of fraud. They want to make an impression that they will be able collect from the tax pater on the charge of fraude. \\
\hline
ALL & During an arrest, tax agents want to make an impression on who do some tax agents want \\
\hline
KA & The proper form of the question is ``Who do some tax agents wantto make and impression on during a arrest?". \\
\hline
SAR & I meant Who do some tax agent want to make and impression upon during an arrest? \\
\hline
\hline
\multicolumn{2}{c}{Case 2} \\
\hline
Instruction & Given an entity, a before event, an after event, and an attribute related to the entity, generate a sentence as output. Your sentence should show the changes in the attribute of the entity. \\
\hline
Input & entity: rope, before: separate from ball, after: through slot in ball, attr: location \\
\hline
Ground Truth & The location of the rope was separate from the ball before and through a slot in the ball afterwards. \\
\hline
\hline
SFT & Rope is strong because it can be wrapped around many things, including a ball. Rope can be wrapped around a ball, and this ability to be wrapped around things makes rope stronger. \\
\hline
KD & Rope is stronger when it is through the slot in the ball \\
\hline
GKD & rope: separation from ball, after: slot in ball, attr: location \\
\hline
ALL & Rope separates from ball.\textbackslash nRope through slot in ball.\textbackslash nAttr location changes. \\
\hline
KA & The rope has changed location. It was once separated from the ball. Now it is through the slot in the ball. \\
\hline
SAR & The rope was separated from the ball. The rope was through the slot in the ball. \\
\hline
\end{tabular}
\caption[Response examples on the S-NI dataset for student models distilled from MoE teacher.]{Response examples on the S-NI dataset for student models distilled from MoE teacher. Response examples from the S-NI dataset demonstrate that student models trained using our proposed MoE-specific KD methodologies (KA, SAR) follow instructions more accurately.}
\label{tab:moe_qualitative}
\vspace{-0.05in}
\end{table*}

Table~\ref{tab:ablation_lambda} shows the performance for different values of $\lambda$, which represent the probability of sampling experts.
In this experiment, we use the Llama-MoE-3.5B (4/16) model as a teacher and fix the value $M$, the number of augmented samples, as 2.
The result indicates that too large $\lambda$ leads to performance degradation.
This result is similar to the pattern observed in Figure~\ref{fig:effect_of_M}, likely due to the analogous reason.
In other words, the proper value of $\lambda$ generally makes augmentation helpful, whereas the excessive value of $\lambda$ compromises the knowledge.

\section{Qualitative Results}
\label{sec:qualitative_results}

For the qualitative results, we present samples generated by student models trained using various methods. The samples are drawn from the S-NI dataset and utilize LLaMA-MoE-3.5B (4/16) as the teacher model, with Sheared-LLaMA-1.3B employed as the student model. Results are shown in Table~\ref{tab:moe_qualitative}. It is shown that our proposed methods generate responses most similar to the ground truth.

\end{document}